\documentclass{article} 
\usepackage{iclr2026_conference,times}

\usepackage{amsmath,amsfonts,bm}

\def\eqref#1{equation~\ref{#1}}

\def\1{\bm{1}}

\DeclareMathAlphabet{\mathsfit}{\encodingdefault}{\sfdefault}{m}{sl}
\SetMathAlphabet{\mathsfit}{bold}{\encodingdefault}{\sfdefault}{bx}{n}

\usepackage{hyperref}
\usepackage{url}
\usepackage{booktabs}
\usepackage{array}
\usepackage{ragged2e}
\usepackage{caption}
\usepackage{graphicx}

\usepackage{amsmath,amssymb,amsthm}

\title{What Does an LLM Learn from Reinforcement Learning? A Mechanistic Interpretability Perspective with Fixed-SAE Track}

\author{Lingheng Du\thanks{Equal contribution.} \\
  Peking University \\
  \texttt{lingheng@stu.pku.edu.cn} \\
  \And
  Yiming Tang\footnotemark[1] \\
  National University of Singapore \\
  \texttt{yiming@nus.edu.sg} \\
  \And
  Xufeng Duan \\
  The Chinese University of Hong Kong \\
  \texttt{xufengduan@cuhk.edu.hk} \\
  \And
  Dianbo Liu$^\dagger$ \\
  National University of Singapore \\
  \texttt{dianbo@nus.edu.sg} \\
}

\iclrfinalcopy 

\begin{document}

\maketitle

\begin{abstract}
Reinforcement learning (RL) is widely utilized in large language model training to improve targeted capabilities, yet how RL reshapes a model remains poorly understood. Prior attempts to explain how RL works largely offer behavioral perspectives, leaving open what RL gives a model at the representation level: can RL create genuinely novel features, and which existing features does it enhance or suppress? Recent developments in mechanistic interpretability suggest sparse autoencoders (SAEs) as a promising lens to decompose internal activations into human-interpretable features; however, they cannot be directly applied to tracking change across training. In this work, we introduce Fixed-SAE Track, a framework that trains one shared SAE per considered layer on activations pooled across the base model and all RL checkpoints, holding every feature direction fixed so that representation shifts are rigorously defined through the activations of interpretable SAE latents, including the detection of emerging novel features. Validated across multiple datasets and RL algorithms, we find that RL-induced drift is small, gradual, concept specific, and concentrated in late layers, mainly enhancing the sampling rates of a small set of ladder tokens, formatting scaffolding such as step breaks and answer delimiters, rather than reshaping problem content. Steering these features into the base model recovers around 80\% of RL's performance gain, suggesting that RL primarily elicits capabilities the model already possesses, much as steering does. We further design a synthetic benchmark with features known by construction to test whether RL can instill genuinely novel features. We believe Fixed-SAE Track provides a principled approach to tracking representation shifts and offers representational evidence for understanding how reinforcement learning changes the inner representation of LLMs.
\end{abstract}

\section{Introduction}

Reinforcement learning (RL), with roots in trial-and-error learning and optimal control \citep{sutton1998reinforcement}, is a paradigm in which an agent learns optimal behaviors through interaction with its environment, guided by reward feedback. It has driven landmark successes in domains such as game playing \citep{mnih2015human, silver2016mastering} and robotic control \citep{kober2013reinforcement}. With the surge of large language models (LLMs) \citep{zou2026fml}, RL has been coupled with human feedback (RLHF) to align model outputs with human preferences \citep{ouyang2022traininglanguagemodelsfollow} and with verifiable rewards (RLVR) to enhance targeted capabilities such as mathematical and code reasoning \citep{shao2024deepseekmathpushinglimitsmathematical, wen2025reinforcementlearningverifiablerewards}. Across these settings RL reliably improves the metric of interest, and it is now a routine stage of modern post-training pipelines \citep{Guo_2025,grattafiori2024llama3herdmodels}.

Despite this widespread adoption, how RL actually reshapes a model remains limitedly understood, and existing accounts offer seemingly contradictory perspectives. Some find that RL merely sharpens the sampling distribution over solutions the base model can already produce \citep{yue2025doesreinforcementlearningreally}, that even spurious rewards improve accuracy by eliciting pretraining priors \citep{shao2026spuriousrewardsrethinkingtraining}, or that RL can even collapse the base model's capability boundary \citep{dong2026rlpluscounteringcapabilityboundary}; others report that RL genuinely incentivize correct reasoning \citep{wen2025reinforcementlearningverifiablerewards} and that prolonged RL can expand the reasoning boundary \citep{liu2025prorlprolongedreinforcementlearning}. Recent analyses further refine our understanding of RL: simple REINFORCE-style methods can match or outperform the more complex PPO for RLHF \citep{ahmadian2024basicsrevisitingreinforcestyle}, positive and negative reinforcement shape output diversity in opposite ways \citep{zhu2025surprisingeffectivenessnegativereinforcement}, and RL's effect concentrates on a minority of high-entropy tokens \citep{wang20258020rulehighentropyminority}. These analyses, however, are conducted almost entirely at the behavioral level, by grading model outputs, and do not reveal what actually changes inside the model.

To peek inside model internals, recent developments in mechanistic interpretability provide valuable tools to decode the features encoded in a model's internal representations \citep{rai2025practicalreviewmechanisticinterpretability, bereska2024mechanisticinterpretabilityaisafety, sharkey2025openproblemsmechanisticinterpretability,2025arXiv251106571Z} and provide insightful discoveries about model representations \citep{yu2026modalitygapdrivensubspacealignment,olsson2022incontextlearninginductionheads,saini2026languageoverwritesvisionoveralignment,todd2024functionvectorslargelanguage}. In particular, sparse dictionary learning (SDL) methods such as sparse autoencoders (SAEs) decompose dense, polysemantic activations into a dictionary of sparse, human-interpretable features \citep{bricken2023monosemanticity, cunningham2023sparseautoencodershighlyinterpretable, gao2024scalingevaluatingsparseautoencoders}. This naturally leads us to ask: from the perspective of a model's representation, what does RL give a model? Can RL training create genuinely novel features, and which existing features does it enhance or suppress?

However, although SDL methods are highly advanced for analyzing static representations, they cannot be directly utilized for the mechanistic analysis of multiple representations. Crosscoders provide a way to compare representations across different models \citep{jiralerspong2026crossarchitecturemodeldiffingcrosscoders}, and researchers have developed methods to match features across layers \citep{balagansky2025mechanisticpermutabilitymatchfeatures} and to track the flow of features through a model \citep{laptev2025analyzefeatureflowenhance}. Nevertheless, SAEs still remain poorly suited to analyze representations across training checkpoints: independently trained SAEs assign feature indices arbitrarily, so features are not comparable between checkpoints, and matching them with additional algorithms further complicates the evaluation of representation shifts. SAE Track \citep{xu2025trackingfeaturedynamicsllm} offers an approach by sequentially fine-tuning SAEs along the training trajectory, but it relies on careful per-checkpoint training and, because each dictionary is inherited from the last, is ill-suited for the detection of emerging novel features.

In this work, we propose \textbf{Fixed-SAE Track}, a novel framework that enables human-interpretable tracking of representation shifts across training, and investigate how RL changes the model representation. Our method first trains one shared SAE per considered layer on activations pooled across the base model and all RL checkpoints, then re-encodes every checkpoint through this fixed dictionary, so that each feature direction is held fixed across training. We further provide rigorous definitions that quantify representation shifts through the activations of the shared SAE latents, which are themselves highly interpretable. Applying Fixed-SAE Track to RL post-training for mathematical reasoning, we find that RL-induced representation shifts concentrate in the model's late layers, and that RL mainly enhances the sampling rates of a small set of ladder tokens, formatting scaffolding such as step breaks and answer delimiters, rather than reshaping problem content. Moreover, steering these features into the base model recovers around 80\% of RL's performance gain, suggesting that RL primarily elicits capabilities the model already possesses, much as steering does. To further test whether RL can create genuinely novel features visible in representation space, we design a synthetic benchmark in which the target features are known by construction.

Our contributions are as follows:
\begin{itemize}
    \item \textbf{Method.} We propose Fixed-SAE Track, a novel framework that enables rigorous and interpretable tracking of the representation shifts induced by model training.
    \item \textbf{Findings.} Our experimental results demonstrate that RL-induced drift is small, gradual, and concentrated in late-layer formatting features; and that steering these features recovers around 80\% of RL's gain, indicating RL primarily elicits model priors.
    \item \textbf{Evidences.} We provide comprehensive evidences to support our scientific findings about reinforcement learning, across multiple models and diverse training settings.
\end{itemize}

\section{Related Work}

\subsection{Reinforcement Learning for Large Language Models}
Reinforcement Learning (RL) has emerged as a powerful paradigm in Artificial Intelligence (AI), enabling agents to learn optimal behaviors through trial-and-error interaction with their environments, guided by reward and penalty feedback \citep{ghasemi2025comprehensivesurveyreinforcementlearning,dong2020deep}. It has been widely utilized to align model outputs with human preferences (RLHF; \citealp{ouyang2022traininglanguagemodelsfollow}) or to enhance mathematical and code reasoning with verifiable rewards (RLVR; \citealp{wen2025reinforcementlearningverifiablerewards}). The policy is typically optimized with policy-gradient algorithms such as REINFORCE \citep{hu2025reinforcestabilizingcriticfreepolicy}, RLOO \citep{Kool2019Buy4R}, PPO \citep{schulman2017proximalpolicyoptimizationalgorithms}, and GRPO \citep{shao2024deepseekmathpushinglimitsmathematical}, while DPO folds reward modeling and policy optimization into a single contrastive objective \citep{rafailov2024directpreferenceoptimizationlanguage}. 
While these works establish that RL reliably improves accuracy, whether it expands a model's underlying capability or merely surfaces behaviors the base model already possesses remains contested. Some find that RL sharpens the sampling distribution without extending the reasoning boundary \citep{yue2025doesreinforcementlearningreally}, that even spurious rewards can elicit latent reasoning from pretraining priors \citep{shao2026spuriousrewardsrethinkingtraining}, and that RLVR might collapse the base model's capability boundary \citep{dong2026rlpluscounteringcapabilityboundary}. Others argue the opposite: RL with verifiable rewards genuinely incentivizes correct reasoning \citep{wen2025reinforcementlearningverifiablerewards}, and prolonged RL training can expand the reasoning boundary \citep{liu2025prorlprolongedreinforcementlearning}. Finer-grained analyses attribute these outcomes to algorithmic details, showing that positive reinforcement sharpens the distribution at the cost of diversity while negative reinforcement preserves it \citep{zhu2025surprisingeffectivenessnegativereinforcement}, and that RL's effect concentrates on a minority of high-entropy forking tokens \citep{wang20258020rulehighentropyminority}.

\subsection{Sparse Autoencoders for Mechanistic Interpretability.} 

Driven largely by AI safety concerns, mechanistic interpretability aims to reverse-engineer the internal computations of neural networks into human-understandable components \citep{rai2025practicalreviewmechanisticinterpretability, bereska2024mechanisticinterpretabilityaisafety, sharkey2025openproblemsmechanisticinterpretability,2025arXiv251205534T}. In particular, sparse dictionary learning (SDL), and sparse autoencoders (SAEs) have become its central techniques, as they decompose polysemantic model activations into a dictionary of monosemantic features \citep{bricken2023monosemanticity, cunningham2023sparseautoencodershighlyinterpretable}. A variety of variants of SAEs have been further proposed, including TopK, BatchTopK, and Matryoshka SAEs \citep{gao2024scalingevaluatingsparseautoencoders, bussmann2024batchtopksparseautoencoders, bussmann2025learningmultilevelfeaturesmatryoshka, Tang_2026_CVPR}. Large language models are frequently used to automatically interpret, name, and assign meaning to them, turning raw feature directions into human-readable concepts \citep{tang2026humanlikecontentanalysisgenerative, luo2024llmdatasetanalystsubpopulation,luo2023promptengineeringlensoptimal}. Building on these foundations, SAEs and related sparse encoders have been applied across diverse settings, including protein-language models \citep{simon2024interplmdiscoveringinterpretablefeatures}, prompt engineering \citep{saini2026bridgingmechanisticinterpretabilityprompt}, and medical imaging \citep{abdulaal2024xrayworth15features, tang2026cxrlaniclanguagegroundedinterpretableclassifier}. While widely applied to analyze static representations, SAEs have seen limited use in tracking how representations evolve during training \citep{xu2025trackingfeaturedynamicsllm}.

\section{Method}
\label{sec:method}

In this section, we introduce Fixed-SAE Track, a novel framework designed to mechanistically interpret the evolution of LLM representation during reinforcement learning (RL). We first describe how we pool internal model activations across the base model and all subsequent RL checkpoints to create a unified training dataset (Section \ref{subsec:pooling}). Building on this, we detail the training of a single, shared Sparse Autoencoder (SAE) that holds feature directions constant across the entire training trajectory (Section \ref{subsec:fixed_sae}). This fixed dictionary enables us to rigorously quantify representation shifts and feature drift over time (Section \ref{subsec:quantifying_shifts}). To ground these mathematical shifts in human-understandable concepts, we outline our token-level attribution methods for interpreting the SAE latents (Section \ref{subsec:interpretation}). Finally, we present the causal intervention techniques—specifically steering and targeted ablations—used to determine whether these identified features causally drive model behavior or merely correlate with it (Section \ref{subsec:steering}).

\begin{figure}[t]
\centering
\includegraphics[width=\linewidth]{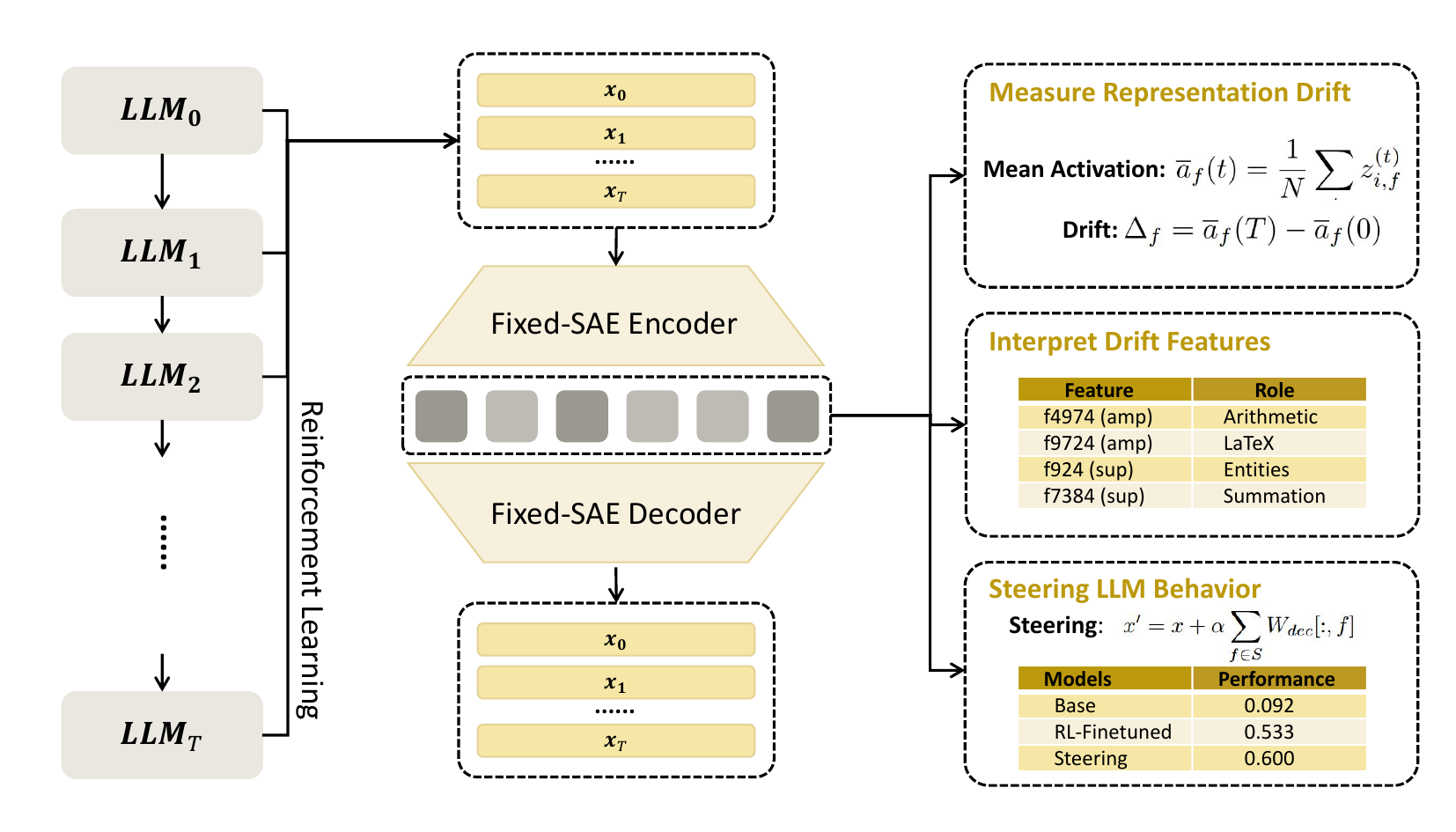}
\caption{\textbf{Overview of Fixed-SAE Track.} Given the base model $\mathrm{LLM}_0$ and its RL checkpoints $\mathrm{LLM}_1,\dots,\mathrm{LLM}_T$, we extract residual-stream activations and train \emph{one} shared SAE on the pooled data, so each feature index denotes the same direction at every training step. Re-encoding every checkpoint through this fixed dictionary supports three downstream tasks: computing drift \emph{metrics}, \emph{interpretation} of the shifted features, and causal \emph{steering} interventions.}
\label{fig:pipeline}
\end{figure}

\subsection{Representation Pooling for RL Checkpoints}
\label{subsec:pooling}
To allow for a valid cross-checkpoint comparison, we first pool activations across the base model and all subsequent RL checkpoints. Specifically, we extract the residual stream vectors $x \in \mathbb{R}^D$ at a late layer (e.g., layer 27) on the final non-pad prompt token. This pooled dataset can be substantial; for example, extracting from multiple checkpoints and multiple layers yields millions of activation vectors. Prior to training the dictionary, we standardize these pooled activations using the global training statistics:
$$ \tilde{x} = (x - \mu) / s $$
where $\mu$ represents the pooled mean and $s$ is the pooled root-mean-square (RMS) scale.

\subsection{Fixed-SAE Training}
\label{subsec:fixed_sae}
Training a separate Sparse Autoencoder (SAE) for each checkpoint produces arbitrary feature indices, rendering cross-checkpoint feature tracking impossible because a feature at step 0 has no relationship to the same index at step 600\citep{2025arXiv251205534T}. We solve this by training a single unified TopK SAE on the pooled, standardized activations \citep{gao2024scalingevaluatingsparseautoencoders}. The TopK SAE computes sparse codes by retaining only the $k$ largest pre-activations:
$$ z = \text{TopK}_k(W_{enc}\tilde{x} + b_{enc}) \in \mathbb{R}_{\ge0}^m $$
$$ \hat{x} = W_{dec}z + b_{dec} $$
After training, this shared dictionary is frozen and used to re-encode the activations of every checkpoint \citep{gao2024scalingevaluatingsparseautoencoders}. As a result, a feature index $f$ corresponds to the exact same direction in activation space—represented by the decoder column $W_{dec}[:,f] \in \mathbb{R}^D$—at every training step.

\subsection{Quantifying Representation Shifts}
\label{subsec:quantifying_shifts}
By re-encoding all checkpoints through the fixed dictionary, we can rigorously define feature drift over time. Letting $z_{i,f}^{(t)}$ denote the code of feature $f$ on example $i$ at checkpoint $t$, we compute the mean activation of the feature as:
$$ \overline{a}_f(t) = \frac{1}{N}\sum_i z_{i,f}^{(t)} $$
We define the drift $\Delta_f$ as the change in mean activation from the base step to the final step $T$:
$$ \Delta_f = \overline{a}_f(T) - \overline{a}_f(0) $$
A positive $\Delta_f$ indicates that the feature was amplified by RL, whereas a negative value indicates that it was suppressed. We additionally monitor the firing rate, which is the fraction of examples on which the feature is active.

\subsection{Interpretation of the Fixed-SAE Latents}
\label{subsec:interpretation}
To map these tracked features to human-interpretable concepts, we attribute them to the specific generated tokens and prompt contexts on which they fire most strongly. We calculate a token-level strong-fire rate over generated tokens:
$$ \rho_f = \frac{1}{T_{gen}} \sum_{tok} \text{1}[z_f > 0.5] $$
This token-level attribution allows us to identify whether a feature corresponds to format scaffolding (such as newlines, section breaks, or LaTeX delimiters) or to actual problem content. Furthermore, we calculate the Spearman correlation between the mean activation $\overline{a}_f(t)$ and the training reward to isolate features that rise and fall monotonically in step with the reward signal.

\subsection{Steering LLMs with Fixed-SAE Latents}
\label{subsec:steering}
Finally, we perform causal interventions to test whether the identified features actively drive model behavior rather than merely correlating with it. To test for causal sufficiency, we steer the base model by uniformly adding the directions of an amplified feature set $S$ into its layer-27 residual stream:
$$ x' = x + \alpha \sum_{f \in S} W_{dec}[:,f] $$
We sweep the steering dose $\alpha$ and assess the model using a two-grader evaluation protocol that independently separates strict format compliance from lenient reasoning correctness. To test for necessity, we ablate target features from the RL-final model by subtracting each feature's specific reconstruction contribution, $\sum_{f \in S} z_f W_{dec}[:,f] \cdot s$, strictly at the tokens where the feature naturally fires.

\section{Results}
\label{sec:results}

We ask a single question: when RL improves mathematical performance, what changes inside the model? Unless noted otherwise, we analyze Qwen2.5-1.5B-Instruct at layer~27 using a fixed TopK SAE trained on activations pooled across the base model and all RL checkpoints. Additional training details, robustness analyses, controls, and qualitative examples are deferred to Appendix~\ref{sec:appendix}.

\subsection{RL induces a small, gradual, and late-layer representation shift}
\label{sec:e1}

\begin{figure}[t]
\centering
\includegraphics[width=0.8\linewidth]{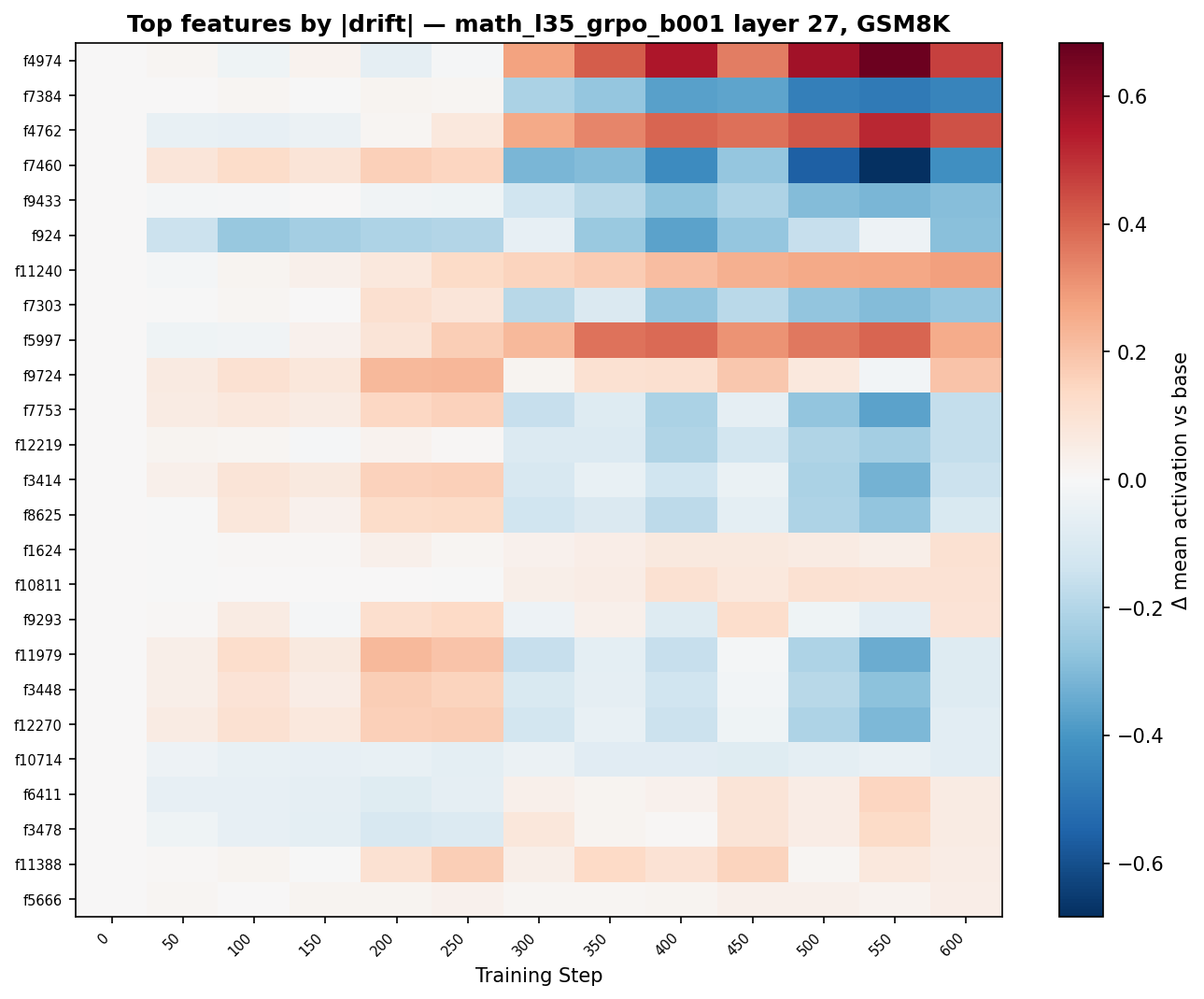}
\caption{\textbf{RL induces concentrated and gradual feature drift.} Change in mean activation relative to the base model, \(\Delta\bar a_f(t)=\bar a_f(t)-\bar a_f(0)\), for the 24 most-changed features across 13 GRPO checkpoints. Only a small subset undergoes substantial amplification or suppression.}
\label{fig:drift}
\end{figure}

We first track the same SAE features across 13 checkpoints of GRPO training. For feature \(f\), we define its drift as \(\Delta_f=\bar a_f(T)-\bar a_f(0)\), where \(\bar a_f(t)\) is its mean activation at checkpoint \(t\). Figure~\ref{fig:drift} shows that the representation evolves smoothly rather than abruptly, with substantial change concentrated in roughly ten features. The largest amplified features include \(f_{4974}\), \(f_{4762}\), \(f_{11240}\), and \(f_{5997}\), while \(f_{7384}\), \(f_{9433}\), and \(f_{924}\) are strongly suppressed. Drift also increases sharply with depth: \(\max_f|\Delta_f|\) rises from \(0.013\) at layer~6 to \(0.060\) at layer~27. Thus, RL does not globally rewrite the representation; it makes a focused edit concentrated near the model output. Active-set turnover, dictionary-health checks, and cross-algorithm reproducibility support the same conclusion (Appendix~\ref{app:drift}--\ref{app:repro}).

\subsection{RL amplifies formatting scaffolds rather than problem content}
\label{sec:e3}

We next ask what the shifted features represent. For each strongly changed feature, we inspect the generated tokens on which it activates most strongly. Table~\ref{tab:attrib} reveals a striking asymmetry: amplified features encode arithmetic operators, LaTeX delimiters, and paragraph or reasoning-step boundaries, whereas suppressed features encode problem entities, units, and intermediate connective text. An independent analysis selecting features by monotonic correlation with reward recovers the same formatting-heavy signature (Appendix~\ref{app:attrib}). The dominant representational effect of short RL is therefore a change in how reasoning is organized and emitted, rather than a wholesale change in problem semantics.

\begin{table}[t]
\centering\footnotesize
\caption{\textbf{Token-level attribution of shifted features.} Amplified features predominantly encode formatting and reasoning scaffolds, whereas suppressed features are more associated with problem content.}
\label{tab:attrib}
\begin{tabular}{lll}
\toprule
Feature & Direction & Interpretation \\
\midrule
\(f_{4974}\)  & amplified & arithmetic operators \\
\(f_{9724}\)  & amplified & LaTeX math-block edges \\
\(f_{4762}\)  & amplified & paragraph / step boundaries \\
\(f_{11240}\) & amplified & new reasoning steps \\
\(f_{924}\)   & suppressed & problem entities / units \\
\(f_{7384}\)  & suppressed & intermediate connective text \\
\bottomrule
\end{tabular}
\end{table}

\subsection{The shifted features causally reproduce RL's dominant extractability gain}
\label{sec:e8}

To distinguish improved reasoning from improved answer presentation, we evaluate every completion three ways: \emph{strict} accuracy requires both the correct answer and the trained answer wrapper, \emph{lenient} accuracy ignores formatting and measures whether the correct answer was reached, and \emph{tag-rate} measures wrapper emission alone. Short RL strongly increases strict accuracy and tag-rate while lenient accuracy changes much less, including under cross-dataset evaluation (Appendix~\ref{app:extract}). This indicates that a large fraction of the early measured improvement comes from making answers that the model can already compute reliably extractable by the grader.

\begin{figure}[t]
\centering
\includegraphics[width=0.8\linewidth]{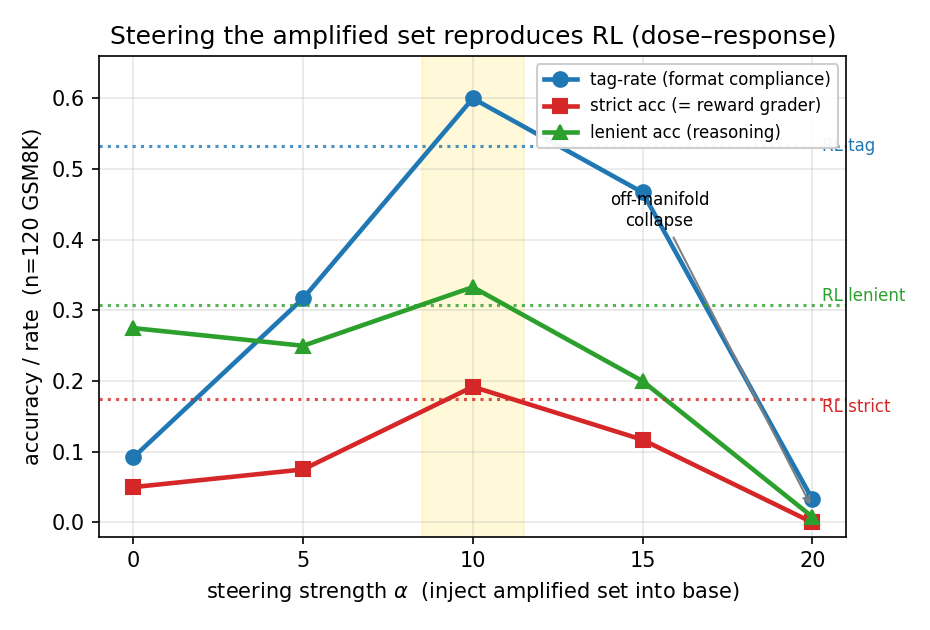}
\caption{\textbf{Steering RL-amplified features reproduces the dominant RL effect.} Adding the amplified feature directions to the base model yields an inverted-U dose response. The peak at \(\alpha=10\) matches the RL-final model within confidence intervals on strict accuracy, lenient accuracy, and tag-rate.}
\label{fig:steer}
\end{figure}

We test causality by adding the decoder directions of the amplified features to the layer-27 residual stream of the base model. Figure~\ref{fig:steer} shows a clear dose response. At \(\alpha=10\), strict accuracy rises from \(0.050\) to \(0.192\) and tag-rate from \(0.092\) to \(0.600\), closely matching the RL-final model (\(0.175\) strict and \(0.533\) tag-rate), whereas lenient accuracy changes only modestly. Random-direction steering does not exhibit the same peak. Moreover, steering directly increases the operator rate and decreases the content-word rate in generated text. The shifted SAE features therefore provide a sufficient causal handle on RL's dominant format/extractability effect.

\subsection{RL mainly elicits existing capability, while sustained training can add reasoning}
\label{sec:elicitation}

The predominance of formatting in short RL does not imply that RL can never improve reasoning. When we restart from the apparent training plateau and continue for 2000 steps with a constant learning rate, lenient accuracy rises from \(0.300\) to \(0.408\), while strict accuracy rises to \(0.408\) and tag-rate to \(0.983\). Thus, sustained training produces a modest genuine reasoning gain in addition to the much larger formatting effect. Temporal analysis further shows that formatting changes occur first: approximately \(86\%\) of total feature amplification is completed within the first 50 steps, while reasoning-associated improvements accumulate more slowly (Appendix~\ref{app:timing}).

We then ask whether the problems solved after RL were genuinely inaccessible to the base model. Using Qwen2.5-Math-1.5B, we isolate 29 GSM8K problems that are wrong under greedy base decoding but correct after RL and sample the base model 64 times on each. Table~\ref{tab:passn} shows that base pass@\(k\) rises from \(0.235\) at \(k=1\) to \(1.00\) at \(k=64\): all \(29/29\) RL-fixed problems are already solvable by the base model under sampling. In this regime, RL therefore primarily increases the probability of successful trajectories already present in the base distribution rather than creating solutions absent from it.

\begin{table}[t]
\centering\footnotesize
\caption{\textbf{Base-model pass@\(k\) on RL-fixed problems.} Among the 29 GSM8K problems that change from base-wrong to RL-right, all are solved by the base model within 64 samples; \(0/29\) are never solved.}
\label{tab:passn}
\begin{tabular}{lcccccc}
\toprule
\(k\) & 1 & 4 & 8 & 16 & 32 & 64 \\
\midrule
base pass@\(k\) & 0.235 & 0.60 & 0.78 & 0.91 & 0.97 & 1.00 \\
\bottomrule
\end{tabular}
\end{table}

This elicitation picture has a mechanistic qualification. We identify a small arithmetic-reasoning feature set \(R\) whose ablation substantially reduces mathematical accuracy while matched format and random controls have negligible effects. Yet directly injecting \(R\) into the base model recovers only a small fraction of RL's gain. Reasoning therefore appears to depend on a dynamically sustained computation rather than a static vector that can simply be pasted into the residual stream. Full construction, causal controls, task specificity, and intervention results are provided in Appendix~\ref{app:R}. Taken together, our results support a qualified account: short RL mainly installs and stabilizes scaffolds that elicit latent capability, while sufficiently sustained RL and models with sufficient headroom can additionally acquire modest genuine reasoning improvements.

\section{Conclusion}
\label{sec:conclusion}
In this work, we introduce Fixed-SAE Track, a framework that trains one shared SAE per layer on activations pooled across the base model and all RL checkpoints, enabling rigorous, interpretable tracking of representation shifts during training. Applying it to RL post-training for mathematical reasoning, we find that RL-induced drift is small, gradual, and concentrated in a handful of late-layer features encoding format scaffolding rather than problem content, and that steering these features into the base model recovers most of RL's accuracy gain --- indicating that RL primarily elicits capabilities the model already possesses. A reasoning feature substrate does exist and is causally necessary, yet it cannot be injected as a static direction, and genuine reasoning gains emerge only when the model has headroom. We believe Fixed-SAE Track offers a principled lens for tracking representation dynamics across training, and we hope it facilitates future mechanistic studies of post-training at larger scales.

\subsubsection*{Acknowledgments}

\bibliography{iclr2026_conference}
\bibliographystyle{iclr2026_conference}

\appendix
\section{Experimental Details and Additional Results}
\label{sec:appendix}

This appendix provides the experimental details and supporting analyses omitted from the main text for space: full training and evaluation protocols, representation-drift controls, cross-algorithm reproducibility, additional feature attribution, the complete format-versus-reasoning decomposition, sustained-training analyses, the construction and causal characterization of the reasoning set \(R\), scale and SAE robustness, failure-mode analyses, and qualitative examples.

\subsection{Experimental and evaluation details}
\label{app:setup}

\paragraph{Models and RL training.} Our primary model is Qwen2.5-1.5B-Instruct (\(D=1536\), 28 layers), analyzed at layer~27. We additionally study Qwen2.5-3B at layer~34, Qwen2.5-7B at layer~27 with LoRA, and Qwen2.5-Math-1.5B BASE/Instruct for the headroom analyses. We train with GRPO, REINFORCE, RLOO, and a stronger GRPO configuration. The binary reward requires the answer to appear inside \verb|<answer>...</answer>| or \verb|\boxed{}| and to be numerically correct. Checkpoints are saved every 50 steps; the primary trajectory contains 13 checkpoints from steps 0--600.

\paragraph{Fixed-SAE training.} The primary dictionary is a TopK SAE with \(k=32\) and expansion factor \(8\), giving \(m=12288\) features for the 1.5B model. Activations are RMS-normalized using statistics computed from the pooled base-plus-checkpoint activation set. After training, the dictionary is frozen and reused for every checkpoint, so decoder column \(W_{\mathrm{dec}}[:,f]\) denotes the same feature direction throughout training.

\paragraph{Two-grader protocol.} We distinguish three quantities. \textbf{Strict} accuracy requires both a correct numerical answer and the trained answer wrapper and therefore matches the RL reward. \textbf{Lenient} accuracy ignores formatting and extracts the answer using \verb|\boxed{}| \(>\) answer tag \(>\) final numerical match, measuring format-agnostic correctness. \textbf{Tag-rate} measures the fraction of generations emitting a valid wrapper irrespective of correctness. Empirically, \(\mathrm{strict}\approx\mathrm{lenient}\times\mathrm{tag\text{-}rate}\): across the steering conditions the absolute decomposition residual is at most \(0.025\).

\begin{table}[h]
\centering\footnotesize
\caption{\textbf{Validation of the two-grader decomposition.} The residual \(\mathrm{strict}-\mathrm{lenient}\times\mathrm{tag\text{-}rate}\) remains within \(0.025\) absolute.}
\label{tab:app-residual}
\begin{tabular}{lccc}
\toprule
Condition & strict & lenient\(\times\)tag-rate & residual \\
\midrule
base & 0.050 & 0.025 & \(+0.025\) \\
\(\alpha=5\) & 0.075 & 0.079 & \(-0.004\) \\
\(\alpha=10\) & 0.192 & 0.200 & \(-0.008\) \\
\(\alpha=15\) & 0.117 & 0.093 & \(+0.023\) \\
\(\alpha=20\) & 0.000 & 0.000 & \(-0.000\) \\
RL-final & 0.175 & 0.164 & \(+0.011\) \\
\bottomrule
\end{tabular}
\end{table}

\subsection{Additional representation-drift analyses}
\label{app:drift}

The active feature set changes continuously rather than discontinuously during RL. Its Jaccard overlap with the base representation decreases monotonically from \(1.0\) to \(0.716\), corresponding to approximately \(28\%\) turnover by the final checkpoint. At the same time, \(L_0\equiv32\) throughout training and reconstruction FVU changes only mildly from \(0.521\) to \(0.539\), indicating that the observed drift is not caused by dictionary degeneration.

\begin{figure}[h]
\centering
\includegraphics[width=0.55\linewidth]{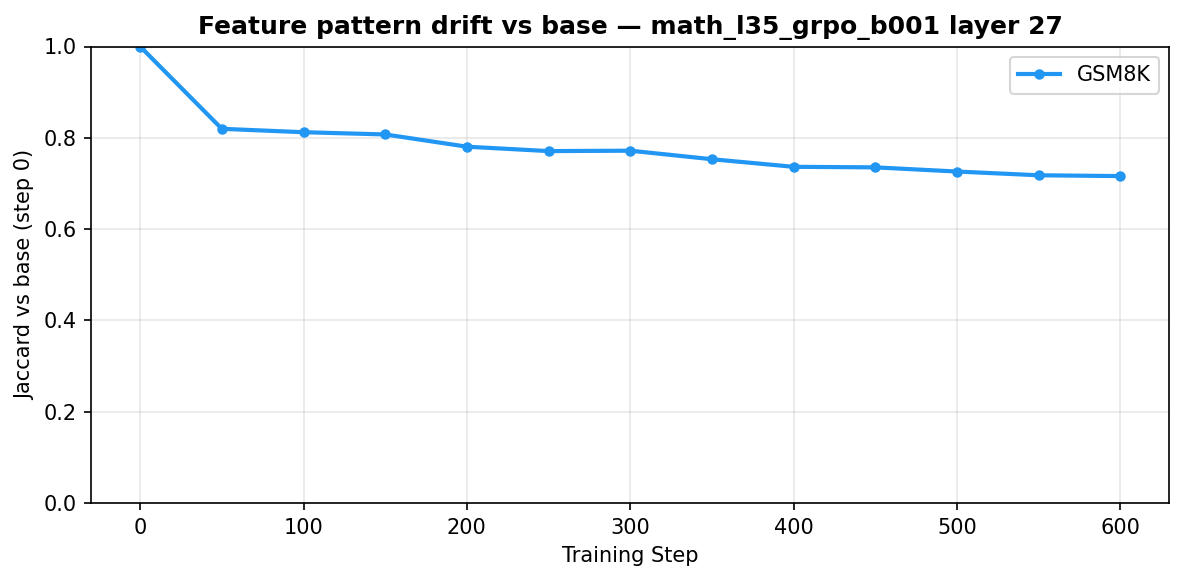}
\caption{\textbf{Active-set turnover.} Jaccard overlap of the active feature set with the base model decreases smoothly from \(1.0\) to \(0.716\) over RL training.}
\label{fig:drift_jaccard}
\end{figure}

We additionally train fixed SAEs independently at several model depths. The maximum feature drift grows monotonically toward the output, from \(0.013\) at layers 6 and 13 to \(0.060\) at layer~27.

\begin{table}[h]
\centering\footnotesize
\caption{\textbf{Layer locus of representation drift.} Maximum feature drift increases strongly toward the final layers.}
\label{tab:layerlocus}
\begin{tabular}{lccccc}
\toprule
Layer \(\ell\) & 6 & 13 & 20 & 24 & 27 \\
\midrule
\(\max_f|\Delta_f|\) & 0.013 & 0.013 & 0.017 & 0.032 & \textbf{0.060} \\
\bottomrule
\end{tabular}
\end{table}

\subsection{Cross-run and cross-algorithm reproducibility}
\label{app:repro}

We compare drift across GRPO on MATH-hard, GRPO on mixed GSM8K+MATH, REINFORCE, RLOO, and strong-GRPO. The headline features preserve their direction across the principal runs, although their magnitudes vary with training strength. Across all four algorithm families the mean off-diagonal Spearman correlation between full drift vectors is \(0.42\), with all pairwise correlations positive. Suppression is more reproducible than amplification: GRPO versus REINFORCE yields top-\(K\) Jaccard \(0.61\) for suppressed features and \(0.28\) for amplified features.

\begin{table}[h]
\centering\footnotesize
\caption{\textbf{Reproducibility of headline feature drift.} Signs and approximate ordering are preserved across datasets and algorithms.}
\label{tab:repro}
\begin{tabular}{lccc}
\toprule
\(\Delta_f\) & GRPO-hard & GRPO-mixed & REINFORCE \\
\midrule
\(f_{4974}\) (amp)  & \(+0.47\) & \(+0.81\) & \(+0.16\) \\
\(f_{5997}\) (amp)  & \(+0.25\) & \(+0.51\) & \(+0.21\) \\
\(f_{9724}\) (amp)  & \(+0.20\) & \(+0.52\) & \(+0.13\) \\
\(f_{11240}\) (amp) & \(+0.28\) & \(+0.63\) & \(+0.11\) \\
\(f_{924}\) (sup)   & \(-0.29\) & \(-0.38\) & \(-0.55\) \\
\bottomrule
\end{tabular}
\end{table}

\begin{figure}[h]
\centering
\includegraphics[width=0.55\linewidth]{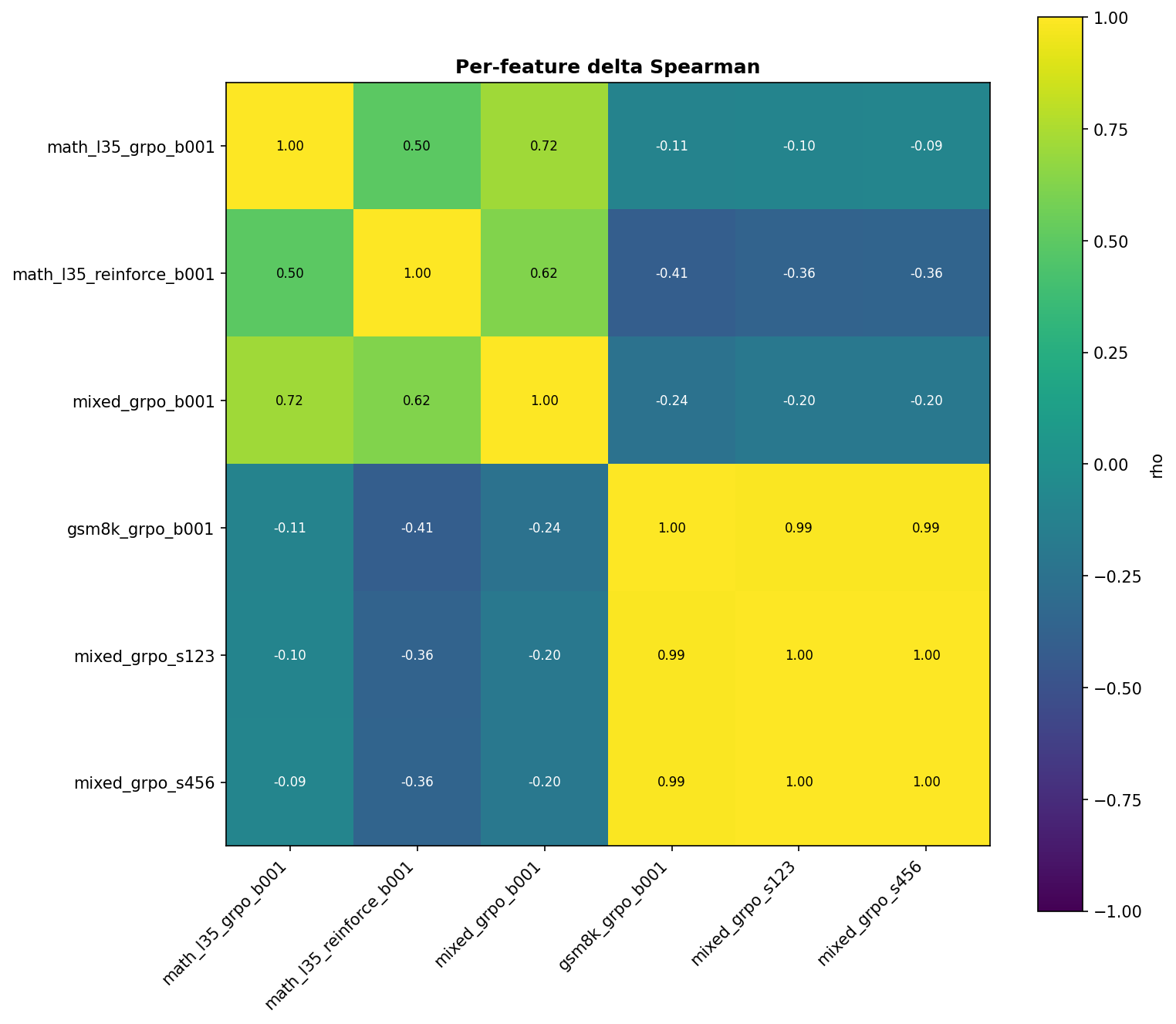}
\caption{\textbf{Cross-algorithm reproducibility.} Pairwise Spearman correlation between feature-drift vectors across four RL algorithms. The mean off-diagonal correlation is \(0.42\), and all pairs are positive.}
\label{fig:repro}
\end{figure}

\subsection{Additional feature-attribution evidence}
\label{app:attrib}

As an independent alternative to ranking features by endpoint drift, we correlate each feature's mean activation trajectory with training reward and retain robust features with \(|\rho_f^{\mathrm{rew}}|\ge0.8\). The resulting amplified features are again dominated by paragraph boundaries, newlines, markdown, and reasoning-step delimiters. For example, \(f_{1213}\) correlates \(+0.93\) with reward and fires on paragraph breaks, while \(f_{10772}\) and \(f_{8439}\) each correlate \(+0.88\) and fire on newline and section/markdown structure. The step-boundary features \(f_{11240}\) and \(f_{4762}\) likewise correlate \(+0.92\) and \(+0.84\), respectively.

\begin{figure}[h]
\centering
\includegraphics[width=0.6\linewidth]{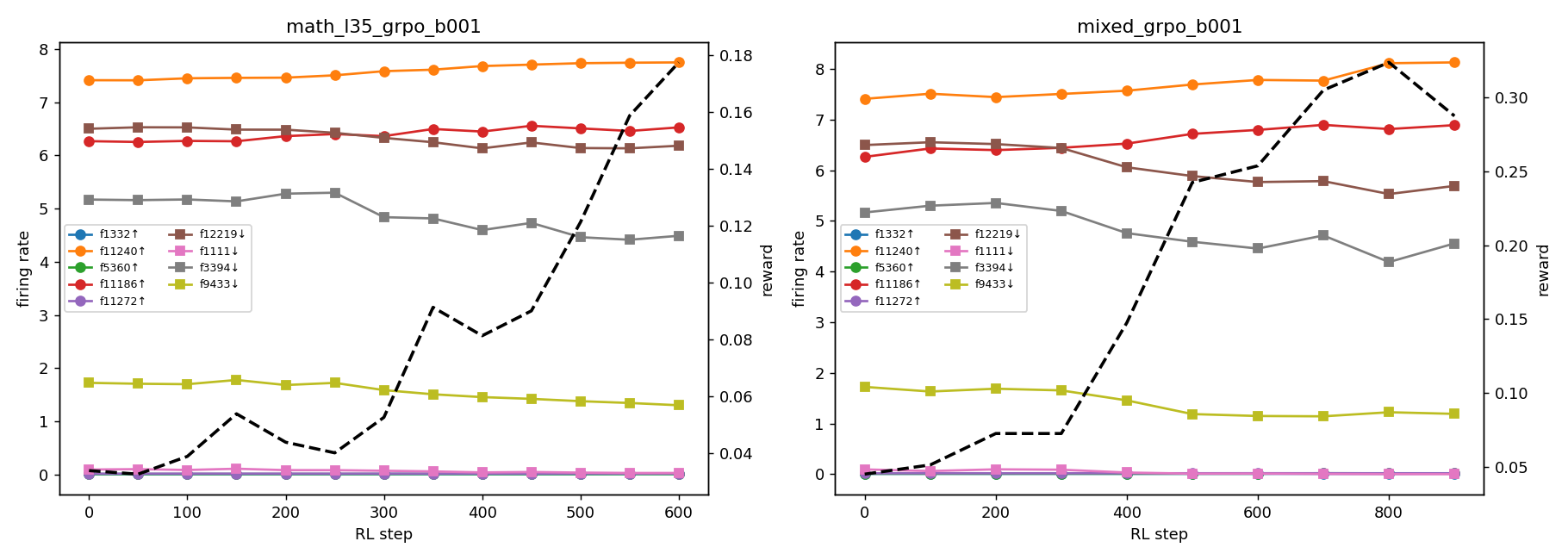}
\caption{\textbf{Reward-monotonic features.} Mean activation trajectories of features whose strength changes monotonically with training reward. Amplified features are predominantly formatting and reasoning-step scaffolds.}
\label{fig:monotonic}
\end{figure}

\subsection{Format-versus-reasoning decomposition and steering controls}
\label{app:extract}

Cross-validation across independently trained RL models confirms that short RL predominantly improves answer extractability. GRPO-trained models strongly increase strict accuracy and tag-rate on both GSM8K and MATH-500, whereas lenient accuracy changes only modestly. Notably, GSM8K-trained GRPO increases MATH-500 strict accuracy by \(+0.13\) while lenient accuracy increases by only \(+0.03\), showing that the transferable component is primarily format.

\begin{table}[h]
\centering\footnotesize
\caption{\textbf{Cross-validation of format versus reasoning.} Changes relative to the base model for independently trained RL checkpoints.}
\label{tab:crossval}
\begin{tabular}{lccc c ccc}
\toprule
& \multicolumn{3}{c}{GSM8K} & & \multicolumn{3}{c}{MATH-500} \\
\cmidrule(lr){2-4}\cmidrule(lr){6-8}
RL run & \(\Delta\)strict & \(\Delta\)lenient & tag & & \(\Delta\)strict & \(\Delta\)lenient & tag \\
\midrule
gsm8k\_grpo    & \(+.23\) & \(+.05\) & .86 & & \(+.13\) & \(+.03\) & .36 \\
math\_grpo     & \(+.14\) & \(+.08\) & .51 & & \(+.06\) & \(-.01\) & .29 \\
mathL35\_grpo  & \(+.12\) & \(+.05\) & .56 & & \(+.09\) & \(+.04\) & .32 \\
mixed\_grpo    & \(+.23\) & \(+.10\) & .89 & & \(+.12\) & \(-.01\) & .34 \\
mathL35\_reinf & \(+.02\) & \(+.01\) & .16 & & \(+.04\) & \(-.02\) & .21 \\
\bottomrule
\end{tabular}
\end{table}

Training strict accuracy tracks reward closely during the short-run regime. Because lenient accuracy is comparatively stable while tag-rate increases rapidly, this trajectory largely measures acquisition of the rewarded output format.

\begin{figure}[h]
\centering
\includegraphics[width=0.7\linewidth]{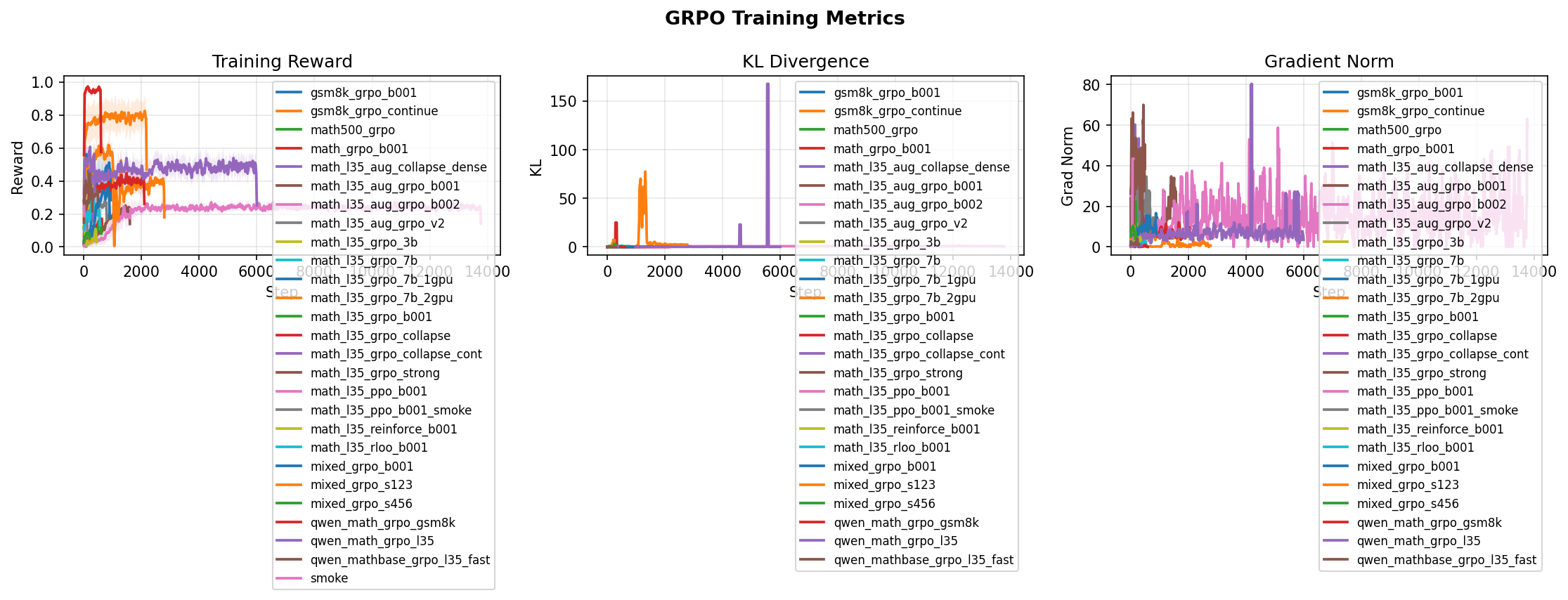}
\caption{\textbf{Training strict accuracy tracks reward.} Strict accuracy increases with training reward on GSM8K and MATH-500 during the short-RL trajectory.}
\label{fig:reward}
\end{figure}

For completeness, Table~\ref{tab:steer-full} reports the numerical steering results underlying Fig.~\ref{fig:steer}. The effect peaks at \(\alpha=10\); larger interventions push the model off-manifold and degrade generation.

\begin{table}[h]
\centering\footnotesize
\caption{\textbf{Full steering dose response.} Wilson 95\% confidence intervals are shown in brackets.}
\label{tab:steer-full}
\begin{tabular}{lccc}
\toprule
Condition & strict & lenient & tag-rate \\
\midrule
base & 0.050 {\scriptsize[.023,.105]} & 0.275 {\scriptsize[.203,.361]} & 0.092 {\scriptsize[.052,.157]} \\
\(+\)steer \(\alpha=5\) & 0.075 {\scriptsize[.040,.136]} & 0.250 {\scriptsize[.181,.334]} & 0.317 {\scriptsize[.240,.404]} \\
\(+\)steer \(\alpha=10\) & \textbf{0.192} {\scriptsize[.131,.271]} & \textbf{0.333} {\scriptsize[.255,.422]} & \textbf{0.600} {\scriptsize[.511,.683]} \\
\(+\)steer \(\alpha=15\) & 0.117 {\scriptsize[.071,.186]} & 0.200 {\scriptsize[.138,.280]} & 0.467 {\scriptsize[.380,.556]} \\
\(+\)steer \(\alpha=20\) & 0.000 {\scriptsize[.000,.031]} & 0.008 {\scriptsize[.001,.046]} & 0.033 {\scriptsize[.013,.083]} \\
\midrule
RL-final & \textbf{0.175} {\scriptsize[.117,.253]} & 0.308 {\scriptsize[.233,.396]} & \textbf{0.533} {\scriptsize[.444,.620]} \\
\bottomrule
\end{tabular}
\end{table}

\subsection{Sustained training and format-then-reasoning timing}
\label{app:timing}

The apparent early training plateau is caused by cosine learning-rate decay rather than an intrinsic capability ceiling. Re-initializing from the plateau checkpoint and continuing for 2000 steps at constant learning rate causes strict accuracy to resume increasing, while lenient accuracy rises from \(0.300\) to \(0.408\). Formatting nevertheless remains the dominant change: tag-rate rises from \(0.108\) to \(0.983\).

\begin{table}[h]
\centering\footnotesize
\caption{\textbf{Sustained constant-LR training.} Longer training yields a substantial format gain together with a modest genuine reasoning gain.}
\label{tab:continue}
\begin{tabular}{lccc}
\toprule
Condition & strict & lenient & tag-rate \\
\midrule
base & 0.050 & 0.300 & 0.108 \\
continue \(+2000\) & 0.408 & 0.408 & 0.983 \\
\bottomrule
\end{tabular}
\end{table}

Feature dynamics exhibit the same temporal ordering. Approximately \(86\%\) of total feature amplification occurs before step~50, dominated by format-scaffold features, whereas reasoning-content features change later. RL therefore installs the output scaffold rapidly and accumulates smaller reasoning improvements more slowly.

\subsection{Reasoning set \(R\): construction and causal characterization}
\label{app:R}

We construct a candidate arithmetic-reasoning feature set \(R\) by intersecting four independently computed criteria: features that are high on rare-but-correct base completions, discriminate difficult mathematical prompts from trivial controls, are amplified across GSM8K/MATH-500/AMC/AIME, and dominate late-generation computation. This yields a compact set including \(R=\{3256,10380,6904,12287,662,10023,\ldots\}\). A separate format set \(F=\{324,876,3762,5260,7912,8977,11997\}\) serves as a placebo control.

Ablating \(R\) from the RL model reduces lenient GSM8K accuracy from \(0.358\) to \(0.242\) and MATH-500 from \(0.208\) to \(0.117\), whereas ablating \(F\) changes accuracy by only \(\pm0.008\). The effect is reproduced across GRPO, REINFORCE, RLOO, and strong-GRPO and remains near zero on MMLU. Its causal scope is narrower than its descriptive activation: although \(R\) also fires on BBH logical-deduction tokens, ablating it leaves BBH logical deduction and ARC-Challenge essentially unchanged. We therefore interpret \(R\) as causally load-bearing for arithmetic computation in the studied models, not as a universal reasoning circuit.

\begin{table}[h]
\centering\footnotesize
\caption{\textbf{Causal characterization of the arithmetic-reasoning set \(R\).} Ablation strongly affects arithmetic reasoning while matched format and non-arithmetic controls remain largely unchanged.}
\label{tab:rcausal}
\begin{tabular}{llp{5.1cm}}
\toprule
Axis & Probe & Effect of ablating \(R\) \\
\midrule
Necessity & GSM8K & \(0.358\to0.242\) \\
           & MATH-500 & \(0.208\to0.117\) \\
           & ablate \(F\) & \(\pm0.008\) \\
\midrule
Specificity & GSM8K & \(-0.300\) \\
            & BBH-logic / ARC & \(\approx0/\approx0\) \\
            & MMLU & \(-0.02\) \\
\midrule
Cross-algo & GRPO/REINF/RLOO/strong & \(-0.34\) to \(-0.37\) on GSM8K \\
           & ablate \(F\) & \(-0.007\), inert \\
\midrule
Random control & size-matched random sets & approximately zero drop \\
\bottomrule
\end{tabular}
\end{table}

\paragraph{Necessity does not imply injectability.} Uniformly adding the directions in \(R\) to the base model does not reproduce RL. A more faithful intervention that clamps \(z_R\) to a target activation peaks at \(T=0.75\) but recovers only \(+0.06\) accuracy compared with approximately \(+0.34\) from RL. Directly boosting the corresponding special-token logits monotonically hurts performance. During generation, the RL model sustains activation of \(R\), whereas the base model's activation progressively decays. These results suggest that \(R\) is better viewed as a dynamically sustained compute state than a static direction.

\subsection{Generality, headroom, and few-shot elicitation}
\label{app:generality}

The qualitative format-amplification signature appears at all three scales studied. The top-amplified features at 1.5B, 3B, and 7B consistently fire on operators, paragraph boundaries, and LaTeX/markdown delimiters. We restrict this claim to descriptive feature attribution: causal ablations are established cleanly at 1.5B but are inconclusive at 3B and 7B.

\begin{table}[h]
\centering\footnotesize
\caption{\textbf{Scale robustness of the format signature.} Top-amplified features belong to similar formatting families across model sizes.}
\label{tab:scale}
\begin{tabular}{ll}
\toprule
Size & Top-amplified features fire on \\
\midrule
1.5B & operators, paragraph breaks, LaTeX delimiters \\
3B & paragraph breaks, indentation, markdown / LaTeX \\
7B & paragraph breaks, markdown, braces, operators \\
\bottomrule
\end{tabular}
\end{table}

The amount of genuine reasoning improvement depends on model headroom. Already tuned Math-Instruct models exhibit little lenient gain, whereas the Math base model gains approximately \(0.10\)--\(0.15\). Public base-to-Instruct comparisons are used only as complementary scale evidence because they conflate SFT and RL.

Few-shot prompting provides an additional elicitation control. Few-shot prompting strongly improves the base model, whereas adding it to the RL model provides no additional gain and can hurt. The feature-level few-shot activation shift also correlates with RL drift (\(\rho=0.46\)), with suppressed-set Jaccard \(0.70\).

\begin{table}[h]
\centering\footnotesize
\caption{\textbf{Few-shot prompting and RL are partially substitutable.} Calibrated accuracy under the official evaluation protocol.}
\label{tab:fewshot}
\begin{tabular}{lcc}
\toprule
Condition & GSM8K & MATH-500 \\
\midrule
base zero-shot & .396 & .360 \\
base few-shot & .748 & .452 \\
RL zero-shot & .760 & .614 \\
RL few-shot & .732 & .450 \\
\bottomrule
\end{tabular}
\end{table}

\subsection{Robustness to SAE architecture}
\label{app:sae}

We train TopK, BatchTopK, Matryoshka, ReLU, and JumpReLU SAEs on a shared activation pool and compare their top-amplified decoder directions using decoder max-cosine. Sparse SAEs agree substantially above the random-unit baseline (\(\approx0.30\) versus \(\approx0.05\)), while the two dense L1 variants agree strongly with each other (\(0.81\)). Cross-family matching is near random because the dense dictionaries are not comparably sparse: TopK has \(L_0=32\), whereas ReLU and JumpReLU have \(L_0\) above \(5000\). We therefore restrict exact feature-identity claims to the sparse SAE family, while the qualitative direction of drift---format up, content down---is robust across architectures.

\begin{figure}[h]
\centering
\includegraphics[width=0.55\linewidth]{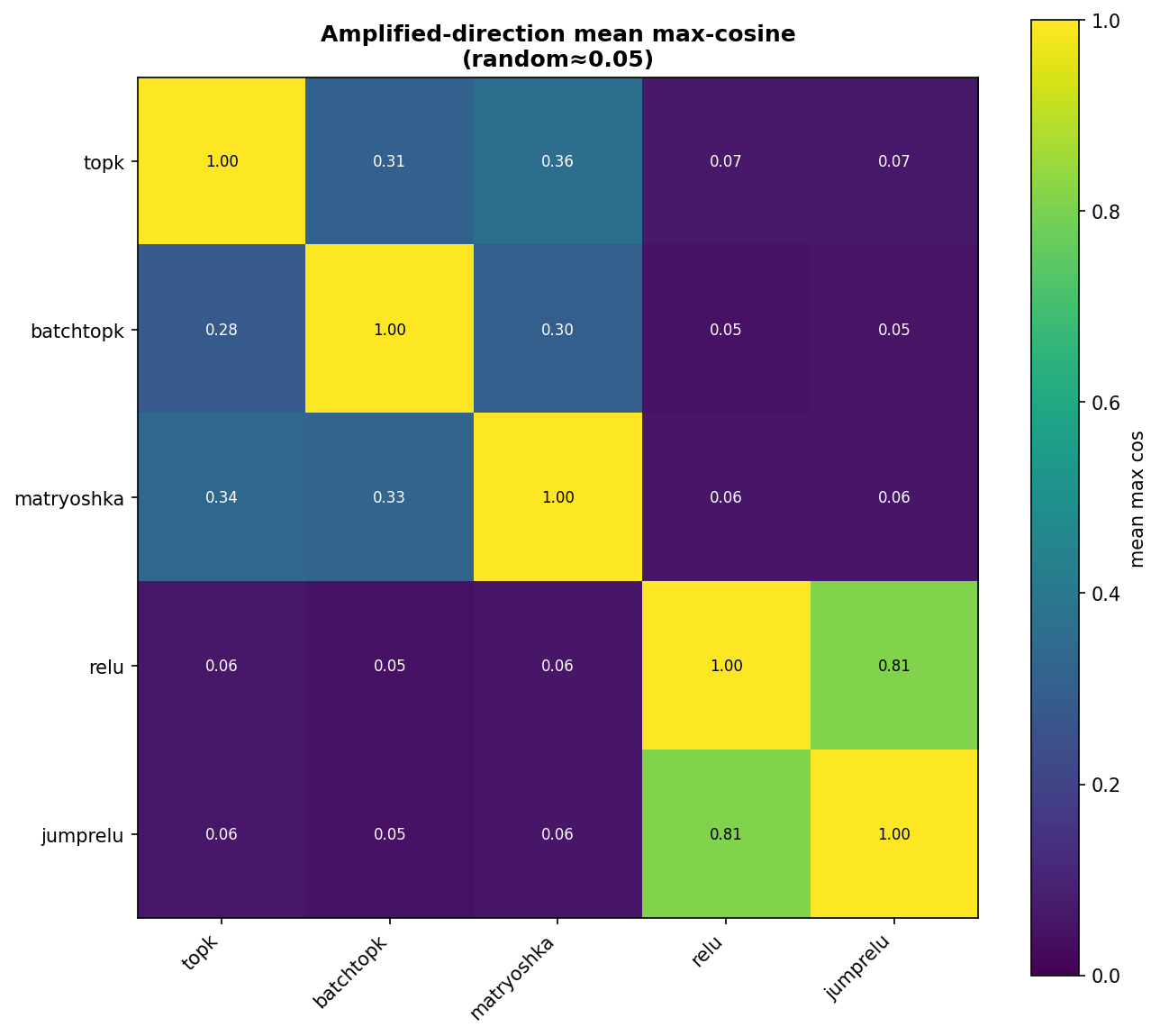}
\caption{\textbf{SAE-architecture robustness.} Decoder max-cosine between top-amplified directions across five SAE families. Sparse architectures agree above the random baseline; dense L1 architectures form a separate family.}
\label{fig:saerobust}
\end{figure}

\subsection{Suppressed features and apparent collapse}
\label{app:collapse}

The features suppressed by RL provide complementary evidence for the elicitation interpretation. In the Math-base representation, strongly suppressed features fire on derailment patterns including malformed Unicode, stray CJK text, code leakage, and runaway newlines. In the Instruct representation, suppressed features are more associated with verbose connective language and heavy markdown. RL therefore suppresses failure and verbosity modes in addition to amplifying formatting scaffolds.

A heavily trained augmented-MATH run appears to ``collapse'' on its training prompts: completion length falls from \(205\) to \(9\) tokens while reward remains high because the model learns to emit a bare answer tag. However, the same checkpoints continue to generate normal-length solutions on held-out GSM8K and MATH-500, and the reasoning set \(R\) remains stable. This behavior is therefore training-distribution reward-hacking brevity rather than global capability collapse.

\begin{figure}[h]
\centering
\includegraphics[width=0.6\linewidth]{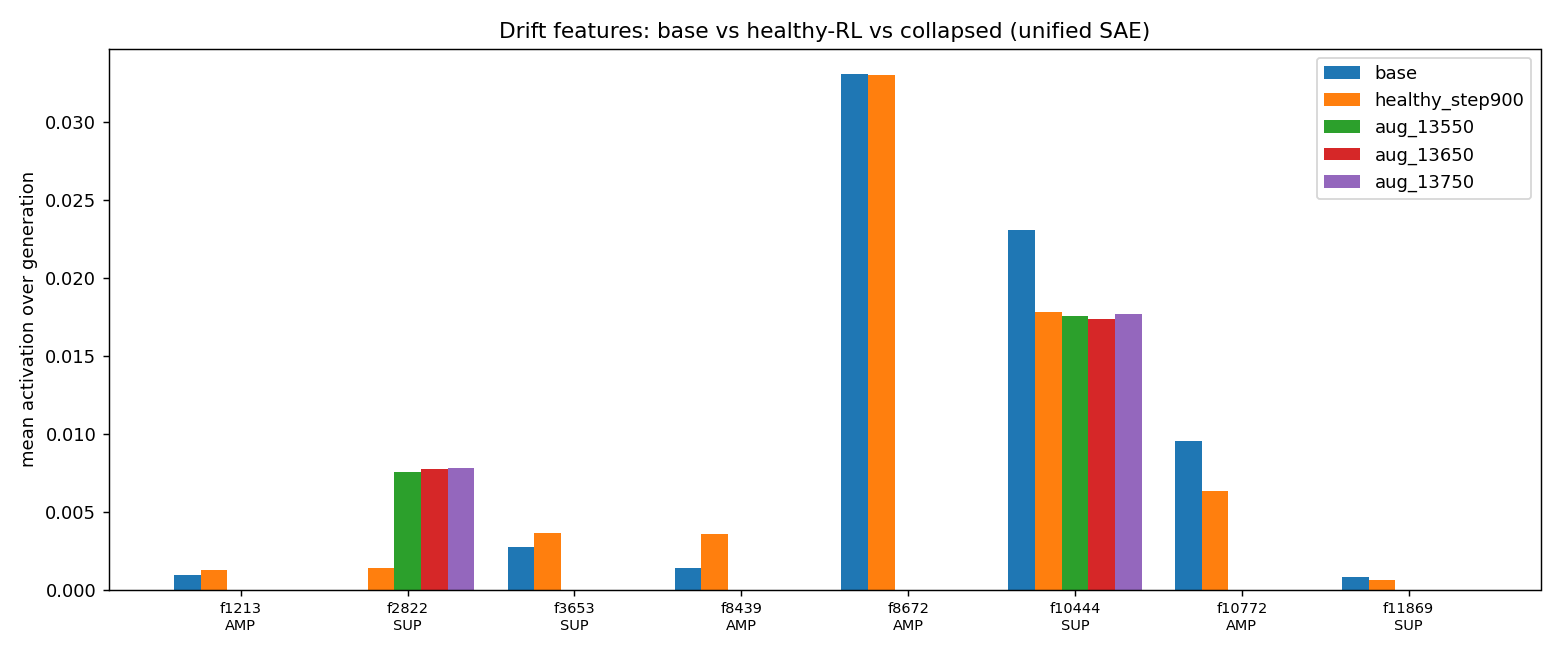}
\caption{\textbf{Re-characterizing apparent collapse.} Training-prompt completion length collapses while reward remains high, but held-out generation remains normal and the reasoning substrate remains active, indicating reward-hacking brevity rather than global capability loss.}
\label{fig:app-collapse}
\end{figure}

\subsection{Qualitative causal examples}
\label{app:cases}

Representative GSM8K generations make the format/extractability mechanism explicit. In the Billy-DVD example, the base model reaches the correct answer \(7\) but omits the answer wrapper and is therefore strict-wrong; the RL model produces essentially the same arithmetic and appends \verb|<answer>7</answer>|. The same pattern occurs for the Gretchen-coins example with answer \(70\).

Ablating the reasoning set \(R\) produces a qualitatively different failure. On a GSM8K problem, the model continues to emit apparently structured reasoning but corrupts operands and intermediate arithmetic, whereas ablating the format control \(F\) changes surface form while preserving the correct answer. On an MMLU knowledge problem, ablating \(R\) is a no-op. These examples support the interpretation that \(R\) affects arithmetic computation rather than merely formatting.

\subsection{Superseded steering analysis}
\label{app:oldsteer}

An earlier steering analysis graded generations only with the lenient metric and used excessively large intervention strengths, leading to the misleading conclusion that amplified-feature steering was ineffective or harmful. This interpretation is superseded by the two-grader dose-response in Fig.~\ref{fig:steer}: lenient accuracy is deliberately insensitive to the formatting improvement produced by steering, while \(\alpha\ge20\) pushes the representation off-manifold.

\begin{figure}[h]
\centering
\includegraphics[width=0.62\linewidth]{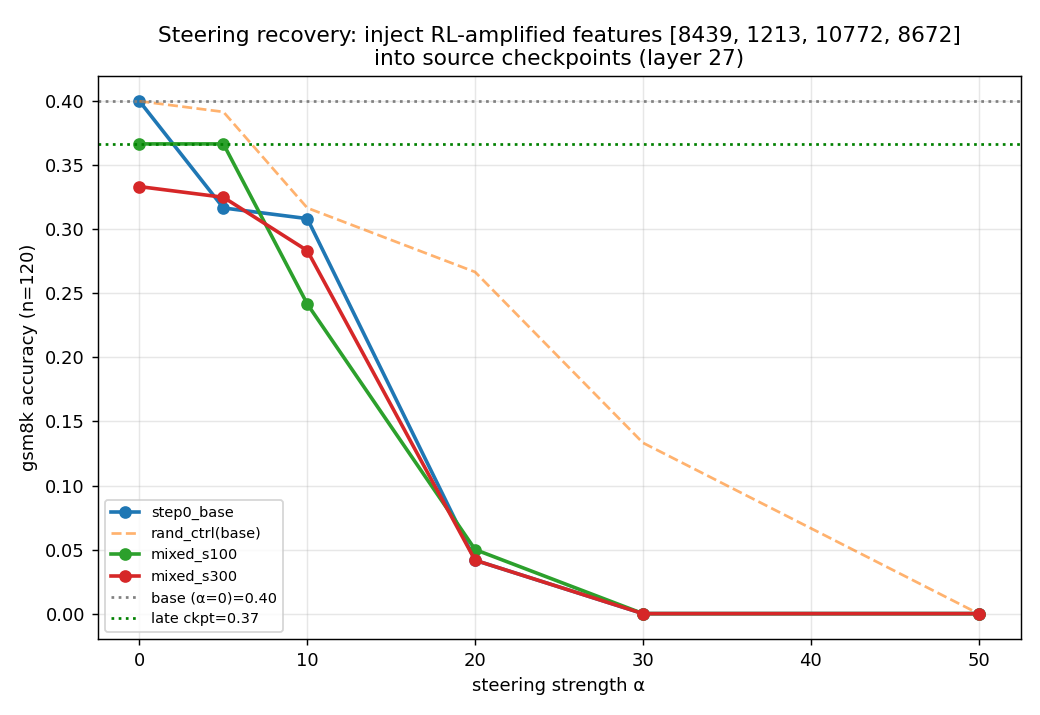}
\caption{\textbf{Superseded steering analysis.} The original lenient-only high-dose analysis obscures the causal format effect and includes off-manifold intervention strengths. It is retained for transparency and should not be interpreted as a primary result.}
\label{fig:app-oldsteer}
\end{figure}

\subsection{Reproducibility notes}
\label{app:repro-notes}

All primary drift measurements use last-token residual-stream activations encoded by the same frozen dictionary. The primary SAE maintains \(L_0\equiv32\) and FVU \(\approx0.52\) throughout training. Exact feature-index identity is interpreted only within the sparse SAE family. Absolute accuracy values reported as calibrated results use the official evaluation protocol rather than the longer and more permissive internal \texttt{cross\_eval} configuration. Pass@\(k\) is computed with the unbiased estimator of \citet{chen2021codex}.

\end{document}